%% file: main.tex
\documentclass[10pt,twocolumn,letterpaper]{article}

\usepackage{cvpr}
\usepackage{times}
\usepackage{epsfig}
\usepackage{graphicx}
\usepackage{amsmath}
\usepackage{amssymb}

\usepackage[pagebackref=true,breaklinks=true,letterpaper=true,colorlinks,bookmarks=false]{hyperref}

\usepackage{color}

\cvprfinalcopy 

\def\cvprPaperID{2570} 

\ifcvprfinal\fi
\begin{document}

\title{Weakly-supervised Visual Grounding of Phrases with Linguistic Structures}

\author{Fanyi Xiao\\
	University of California, Davis\\
	{\tt\small fyxiao@ucdavis.edu}
	\and
	Leonid Sigal\\
	Disney Research\\
	{\tt\small lsigal@disneyresearch.com}
	\and
	Yong Jae Lee\\
	University of California, Davis\\
	{\tt\small yongjaelee@ucdavis.edu}
}

\maketitle

\begin{abstract}
\vspace{-0.1in}
We propose a weakly-supervised approach that takes image-sentence pairs as input
and learns to visually ground (i.e., localize) arbitrary linguistic phrases, in the form of spatial \emph{attention masks}. Specifically, the model is trained with images and their associated image-level captions, without any explicit region-to-phrase correspondence annotations.
To this end, we introduce an end-to-end model which learns visual groundings of phrases with two types of carefully designed loss functions. In addition to the standard discriminative loss, which enforces that attended image regions and phrases are consistently encoded, we propose a novel \emph{structural loss} which makes use of the parse tree structures induced by the sentences. In particular, we ensure complementarity among the attention masks that correspond to sibling noun phrases, and compositionality of attention masks among the children and parent phrases, as defined by the sentence parse tree. We validate the effectiveness of our approach on the Microsoft COCO and Visual Genome datasets.
\end{abstract}

\input{introduction}

\input{relatedwork}

\input{approach}

\input{results}

\input{conclusion}

\vspace{-10pt}
\paragraph{Acknowledgements.} This work was supported in part by GPUs donated by NVIDIA.

{\small
\bibliographystyle{ieee}
\bibliography{strings,mybib,bibs}
}

\end{document}

%% file: introduction.tex
\vspace{-0.08in}
\section{Introduction}
\vspace{-0.02in}

Visual recognition research has made tremendous strides in recent years, achieving unprecedented performance in various tasks including image classification~\cite{He-cvpr2016,Krizhevsky-nips2012, Simonyan-arxiv2014}, object detection~\cite{Girshick-pami2015, Ren-nips2015}, semantic segmentation~\cite{Hariharan-cvpr2015, Long-cvpr2015}, and image captioning~\cite{Chen-cvpr2015,Karpathy-cvpr2015,Xu-arxiv2015}. However, traditional supervised frameworks for these tasks often rely on large datasets with expensive bounding box or pixel-level segmentation annotations.  As the field pushes toward solving larger-scale and more complex problems, obtaining massive annotated datasets is becoming a critical bottleneck.

\begin{figure}[t!]
\centering
\includegraphics[width=0.48\textwidth]{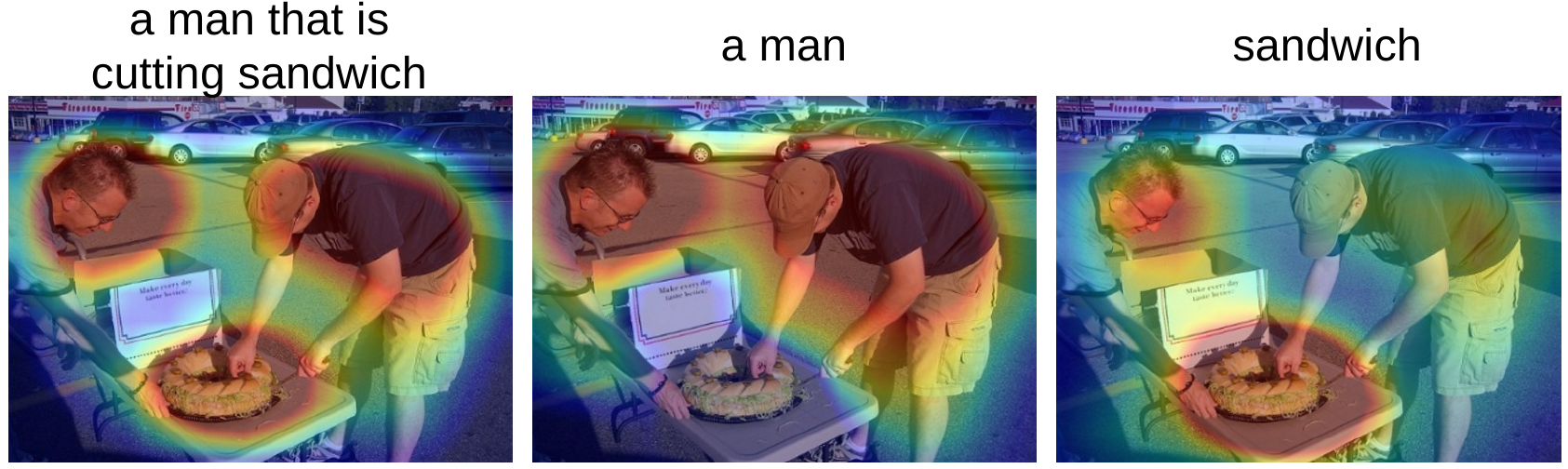}
   \caption{\small{We propose to localize phrases in images, by exploiting linguistic structure. For example, from the phrase ``a man that is cutting sandwich", we can infer that ``a man'' and ``sandwich" should be \emph{exclusive} to each other spatially. At the same time, they should \emph{jointly occupy} the spatial extent of ``a man that is cutting sandwich".  We enforce these structural constraints as part of a novel deep network architecture for weakly-supervised visual grounding of phrases.}}
\label{fig:concept}
\vspace*{-0.1in}
\end{figure}

Weakly-supervised approaches that learn from image-level supervision have been proposed to alleviate the need for expensive and unwieldy annotation.  Most previous work use category tags to train models that can localize objects without any bounding box or segmentation annotations~\cite{cinbis-cvpr2014,fergus-cvpr2003,Kantorov-eccv2016,Karpathy-cvpr2015,Oquab-cvpr2015,Pathak-iccv2015,Krishna-cvpr2016,siva-eccv2012,Song-nips2014,weber-eccv2000}.  While great progress has been made, learning from a list of category tags ignores the rich semantics and structure in natural language that we humans use to describe visual data.  For example, in Fig.~\ref{fig:concept}, a tag-based description would simply list \{man, sandwich, table\} whereas a natural language description might say ``a man that is cutting sandwich on a table''. Importantly, the natural language description provides \emph{structure}, which can benefit a weakly-supervised learning algorithm. For instance, the description implies that ``a man'' and ``sandwich'' occupy \emph{spatially-distinct} regions in the image, and that a visual grounding (localization) of the entire sentence should be the \emph{union} of the groundings of ``a man'' and ``sandwich on a table''. By exploiting these constraints and regularities that are shared between linguistic and visual data, localization in the challenging weakly-supervised setting can be facilitated.

\begin{figure}[t!]
\centering
\includegraphics[width=0.48\textwidth]{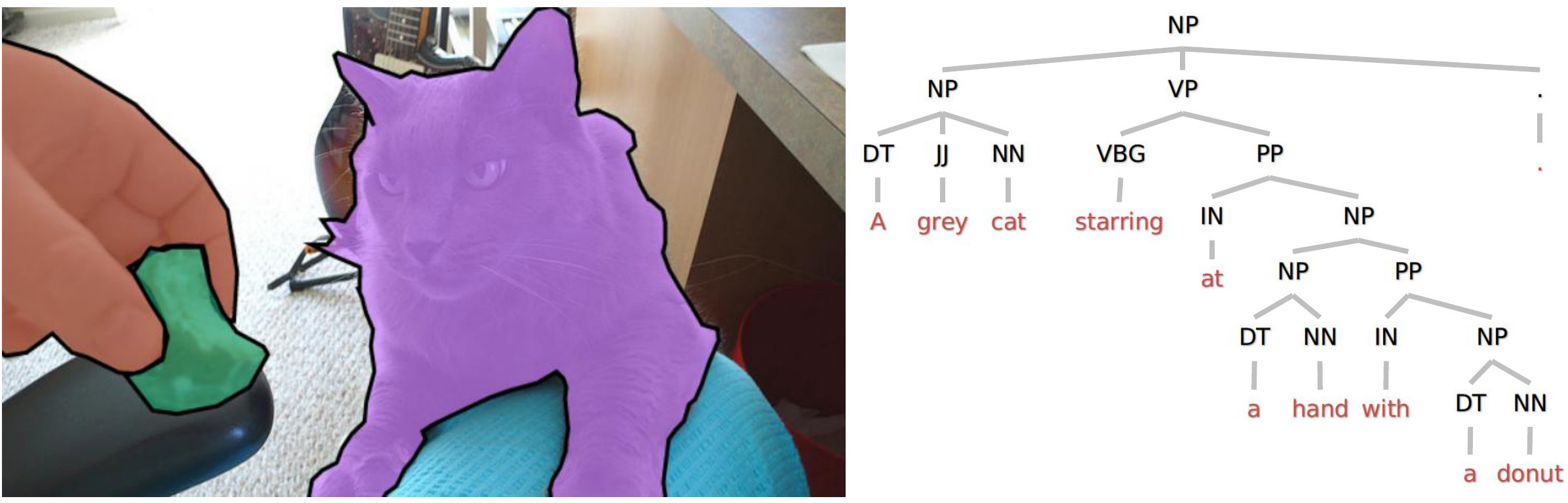}
   \caption{\small{An illustration of the concept.  We exploit structures present in natural language to provide regularities and constraints for grounding free-form language on images. Note that we \textbf{do not} use any ground-truth masks during training.}}
\label{fig:gtexample}
\vspace*{-0.1in}
\end{figure}

In this paper, we propose a weakly-supervised visual localization approach that learns from image-level descriptions (i.e., without any region-to-phrase correspondence annotations).  In particular, we aim to create spatial \emph{attention masks} that produce localizations at the pixel-level.  Our key idea is to utilize the rich structure in a natural language description by transforming it into a hierarchical parse tree of phrases (see Fig.~\ref{fig:gtexample}).  In this way, we can extract two types of linguistic structural constraints for visual grounding: (1) \emph{compositionality} of attention masks among children and their parent phrases (e.g., the mask of ``a hand with a donut" should be the union mask of ``a hand" and ``with a donut"), and (2) \emph{complementarity} among the attention masks between sibling phrases (mask of ``a grey cat" should be spatially-disjoint with that of ``starring at a hand with a donut").  Furthermore, the parse tree representation augments the total amount of supervision and enables learning from linguistic descriptions at various levels, i.e., from words, to phrases, and eventually to the full sentence, for the same image.

Our model is an end-to-end deep neural network that consists of a language encoder, a visual encoder, a semantic projection module, and loss modules.  The language/visual encoder process input phrases/images into vector representations, which are then projected into a common semantic space by the semantic projection module. In addition to the discriminative loss which directly enforces the annotated image-phrase pairs to be close to each other in semantic space, we propose a novel structural loss which enforces the linguistic structural constraints to be satisfied when generating localized attention masks in the images.  Given a test image, our model can generate an attention mask for any arbitrary phrase without needing access to the structures of the linguistic input (e.g., it can localize single word inputs).

Our work is most related to~\cite{Rohrbach-eccv2016}, which learns to localize textural phrases in a weakly-supervised setting, and~\cite{Karpathy-cvpr2015}, which learns to align text to regions for image captioning.  However, unlike our approach, these methods do not explicitly exploit the hierarchical structure in the language description, and instead treat it simply as a sequence of words.  Furthermore, their visual localizations are in the form of bounding boxes, which are insufficient for representing objects with irregular shapes.   Finally, our work is also related to approaches in image captioning~\cite{Mansimov-iclr2016,Xu-arxiv2015} and visual question answering~\cite{Das-emnlp2016,Zhou-arxiv2015}.  While these tasks require localization as a sub-task to be successful, the end goal itself is not localization.  Thus, these approaches do not leverage any structural \emph{localization} constraints as we do.

\vspace{-10pt}
\paragraph{Contributions.} To our knowledge, we are the first to leverage the hierarchical structure of natural language descriptions for weakly-supervised visual localization.  We design a novel deep network with a new \emph{structural loss}, which makes use of the parse tree structures induced by the descriptions.  The structural loss is combined with a discriminative loss, which enforces that attended image regions and phrases are consistently encoded, to produce pixel-level spatial attention masks.  We demonstrate the effectiveness of our approach through localization experiments on the Microsoft COCO and Visual Genome datasets~\cite{Krishna-arxiv2016,Lin-eccv2014}.

%% file: relatedwork.tex
\section{Related work}

\paragraph{Weakly-supervised learning with categorical labels/tags.}
Weakly-supervised visual learning approaches focus on learning granular detectors given only coarse annotations.
This is an extremely useful paradigm since granular annotations (e.g., bounding boxes, segmentations) are much more costly compared to coarse image-level annotations.

Most previous weakly-supervised approaches work with categorical labels/tags. For example, weakly-supervised detection methods aim to train object/attribute detectors with only image-level labels (e.g., whether an object/attribute exists in the image or not) instead of bounding boxes~\cite{Cinbis-pami2016, Deselaers-eccv2010, Pandey-iccv2011, Song-nips2014, berg-eccv2010,wang-cvpr2013b,xiao-iccv2015,krishna-eccv2016}. Despite being a much harder problem, there are also previous efforts in weakly-supervised semantic segmentation, which requires per-pixel predictions~\cite{Pathak-iccv2015, Pinheiro-cvpr2015, Xu-cvpr2014}. Although these methods have demonstrated promising results, the type of annotation used is not very ``natural".  Referring back to Fig.~\ref{fig:gtexample}, it would be somewhat unnatural for someone to tag the image with all depicted objects -- e.g., ``cat, hand, donut".  Instead, it would be much more natural to tag it with a descriptive sentence, similar to what one would expect in a social media post. Assuming this is the case, paring down a sentence to a set of object tags seems to be sub-optimal as the process loses valuable linguistic structure present in the original sentence. Unlike these two lines of work, we perform weakly-supervised learning with \emph{sentence-level} supervision.

\vspace{-10pt}
\paragraph{Vision and language.}
The interplay of vision and language has been studied extensively in recent years, partly because research in both vision and language has matured and has made tremendous progress with the help of deep learning. Given the natural connection between language and vision (e.g., visual language grounding) it is no surprise that multimodal learning that considers both carries significant promise.

Image captioning~\cite{Chen-cvpr2015, Donahue-cvpr2015, Johnson-cvpr2016, Karpathy-cvpr2015, Kiros-arxiv2014, Mansimov-iclr2016, Xu-arxiv2015} has received a great amount of attention in the past two years. Most models adopt an encoder-decoder architecture, in which the encoder encodes information from the image (usually using a CNN) into a hidden state and the decoder then decodes it into a sequence of word tokens (a sentence). The decoder often takes a form of an RNN (e.g., LSTM) and is conditioned on previously generated word tokens.  Visual question answering (VQA) \cite{Antol-iccv2015, Das-emnlp2016, Lu-nips2016} is another popular problem in this space. In VQA, the encoder, in addition to the image, takes a question (often encoded by an RNN) and the decoder decodes the answer.

In both image captioning and VQA, localization of relevant regions helps in generating captions/answers, which motivates many recent models to incorporate \emph{attention mechanisms} to focus on the relevant spatial regions as part of the decoding process~\cite{Lu-nips2016, Karpathy-cvpr2015,Xu-arxiv2015, Das-emnlp2016, Mansimov-iclr2016}. However, since localization is not the final goal, most of these models are either not optimized for it explicitly or simply take off-the-shelf strongly-supervised object detectors for localization.  In contrast, our goal is to perform weakly-supervised {\em localization} directly, and we propose a novel structural loss to facilitate this task.  Other than captioning and VQA, there are also works trying to localize phrases in images~\cite{plummer-iccv2015,wang-cvpr2016,Rohrbach-eccv2016, Wang-eccv2016,wang-cvpr2016,Hu-cvpr2016,Hu-eccv2016}. However, most existing work use strong supervision as the ground-truth correspondence between phrases and \emph{image regions}, which is costly to acquire at scale. While Rohrbach et al.~\cite{Rohrbach-eccv2016} work with weakly-annotated image-phrase pairs, they treat the phrase input as a sequence of tokens, whereas our approach \emph{explicitly} makes use of the structure present in the sentence input.

\vspace{-10pt}
\paragraph{Structure from language.}
We note that we are not the first to utilize linguistic structure for a vision task. In~\cite{Wang-eccv2016}, the authors propose to model ``partial match coreference" relations between phrases. However, the approach requires strong supervision and produces bounding boxes instead of per-pixel predictions. In the VQA approach of~\cite{Andreas-cvpr2016}, the network structure is dynamically constructed using a question; however, it does not directly aim at localization.  To our knowledge, we are the first to leverage the hierarchical structure in natural language descriptions for weakly-supervised visual localization. 

%% file: approach.tex
\section{Approach}\label{approach}

Our goal is to train a model that takes as input a set of weakly-annotated image-sentence pairs (without any explicit region-to-phrase correspondence annotations), and learns to visually ground (i.e., localize) arbitrary linguistic phrases in the form of pixel-level spatial attention masks.  The key idea is to transfer the linguistic structure to the image domain as constraints to guide the model to produce more accurate localizations.  To this end, we propose a novel end-to-end deep network that encodes both the association of phrases and images (with a discriminative loss), as well as the structure of the phrases (with a structural loss).

\subsection{Transforming a sentence into a parse tree}

Given an image and its associated sentence description, we first transform the sentence into a parse tree with an off-the-shelf NLP parser~\cite{Socher-acl2013}, as shown in Fig.~\ref{fig:gtexample}.  In most cases, the structure in the parse tree can also be well-represented in the image (e.g., in Fig.~\ref{fig:gtexample}, ``A grey cat" and ``a hand with a donut" should be exclusive  to each other according to the linguistic structure, which is also true in the visual domain).  We therefore transfer this structure to its corresponding image when visually grounding different nodes in the tree.  In order to ensure that a node corresponds to a region in the image, we only consider nodes that contain at least one noun.  This removes meaningless/ambiguous nodes like ``with", ``and", ``on it", or ``in it".

\subsection{Network architecture}

\begin{figure*}[t!]
\centering
\includegraphics[width=0.95\textwidth]{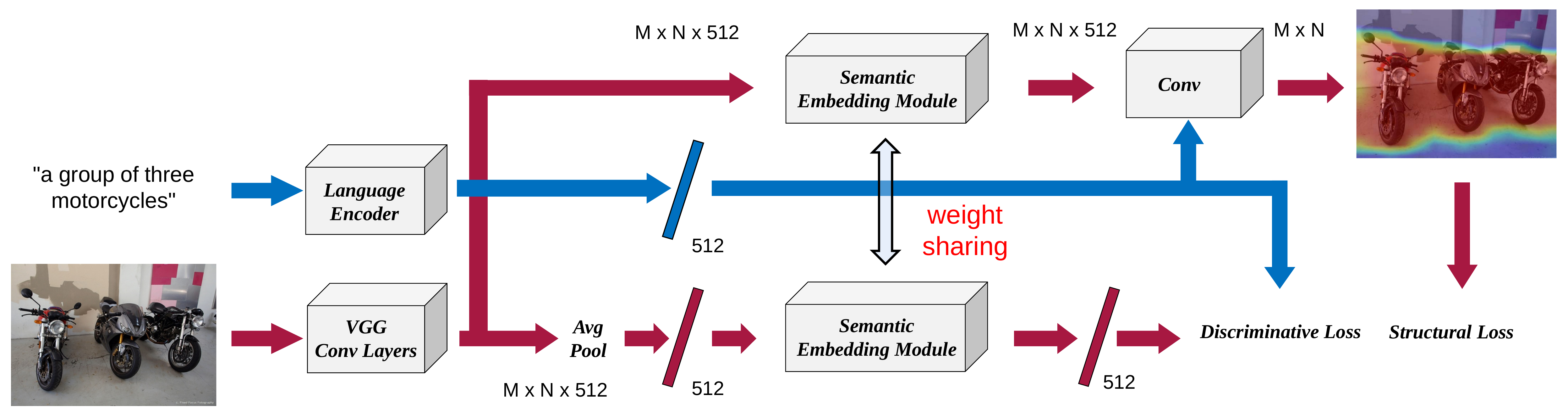}
   \caption{\small{Our architecture consists of 4 submodules: visual encoder, language encoder, semantic embedding module, and loss functions. We adopt the convolutional layers of the VGG network as the visual encoder. For the language encoder, we use a two-layer LSTM network. The output of the language encoder directly lives in the semantic space, whereas the output of the visual encoder is projected into the semantic space by the semantic embedding module, which is a two-layer-perceptron with Dropout inserted in between the layers.  Alongside projecting the visual feature of the full image, the semantic embedding module is also responsible for projecting the feature in each spatial location, to the embedding space: in this case, the output of the visual encoder bypasses average pooling and is directly fed into the embedding module. The embedding modules used for these two purposes share their weights.  After projection, both the full image feature and the spatial feature are matched against the language codes, to generate a matching score and an attention mask, respectively.  The matching score is used to compute the discriminative loss while the structural constraints are enforced onto the attention mask.}}
\vspace{-0.05in}
\label{fig:architecture}
\end{figure*}

The proposed architecture for grounding phrases in an image is shown in Fig.~\ref{fig:architecture}.  There are four parts to the architecture: visual encoder, language encoder, semantic embedding sub-network, and the loss functions.

The visual encoder and language encoder are responsible for encoding the raw input of images and phrases, respectively, into semantic representations.  The semantic embedding sub-network projects the representations from both modalities to a \emph{common semantic space} in which the visual and language data are directly comparable (i.e., enabling one to compute an image-phrase similarity).  In addition to embedding an entire image into the semantic space, the semantic embedding sub-network also projects individual image regions into the common semantic space. To extract these regions a spatial \emph{attention mask} over the image is computed.  Finally, the image-phrase matching scores and attention masks are fed into the discriminative and structural loss modules, respectively, to optimize the network for learning semantics and localization.

In testing, the network takes a phrase as input, and outputs a corresponding attention mask to localize the phrase.

\vspace{-10pt}
\paragraph{Visual encoding.}

We use a convolutional neural network (CNN) to encode the visual content in an image. Specifically, we adopt the VGG-16 network~\cite{Simonyan-arxiv2014} (denoted as VGG), for its high performance and moderate computational cost.  For our use, we remove the fully-connected layers and only keep the convolutional layers ({\tt conv1\_1} through {\tt conv5\_3}), so that we can preserve spatial information for localization.  We initialize the network weights by pre-training on ImageNet~\cite{Deng-cvpr2009}.

\vspace{-10pt}
\paragraph{Language encoding.}

We use a recurrent neural network (RNN) to encode the text descriptions.  The network is able to take as input both short phrases like ``A man" as well as long ones like ``A man riding on the top of an elephant".  To better model long phrases, we adopt LSTM cells~\cite{Hochreiter-neuralcomp1997} in a two-layer RNN, with a Dropout module~\cite{Srivastava-jmlr2014} inserted in between to prevent over-fitting. For a phrase with tokens $\{W_1, W_2, ..., W_T\}$, its representation is computed as the RNN hidden vector at time step $T$.  We pre-train the weights of our language encoder on a combined set of Google's Billion Words ~\cite{Chelba-arxiv2013} dataset and COCO captions in the training ({\tt train2014}) set with the next word prediction task.

\vspace{-10pt}
\paragraph{Joint semantic embedding.}\label{jointsemanticembedding}

We next describe how to project the visual and language representations into a common semantic embedding space.  We directly take the output of the language encoder as the language code in the semantic space. For the output of the visual encoder, i.e., {\tt conv5\_3} map of VGG, we apply the semantic embedding sub-network to obtain the visual code in the semantic space.

Since we want our network to learn to localize the relevant image regions for a given phrase (recall we only have image-level phrase annotations), we feed the {\tt conv5\_3} feature map into a global \emph{average-pooling} layer instead of a max-pooling layer, which is used in the standard VGG network.  As argued in~\cite{Zhou-cvpr2016}, average-pooling better preserves location information since it is forced to localize in order to maximize its response over the relevant image regions.  A 2-layer-perceptron (fully-connected network) follows the average-pooled feature to compute a vector representation for the image.  In order to match the scales of the language and visual codes, we add a batch normalization layer~\cite{Ioffe-nips2015} at the end of the semantic embedding sub-network.

\vspace{-10pt}
\paragraph{Generating spatial attention masks.}\label{attentionmask}

In addition to projecting the {\tt conv5\_3} feature map into a visual code, the semantic embedding sub-network also serves to produce a spatial \emph{attention mask} for each textual phrase.  The attention mask is used both to enforce the structural loss constraints during learning (described in Sec.~\ref{structuralloss}) as well as to produce localizations during testing.  Specifically, we pass individual features at each spatial position of the {\tt conv5\_3} feature map through the network.  The resulting attention mask shows how well the visual feature at each spatial location matches the input textual phrase.

\subsection{Loss for weakly-supervised visual grounding}%

Finally, we introduce the loss functions that we use to induce visual grounding in a weakly-supervised setting. Our architecture is trained end-to-end with two loss terms:

\begin{equation}\label{loss}
L = L_{struct} + L_{disc}.
\end{equation}

The structural loss $L_{struct}$ enforces structure encoded in the text phrases to be satisfied by their respective visual attention masks.  The discriminative loss $L_{disc}$ enforces positive/negative image-phrase pairs to be close/far from each other in the semantic embedding space.

\subsubsection{Structural loss}\label{structuralloss}

We propose to exploit the rich hierarchical structure from the language input to help disambiguate the visual grounding of phrases.  Unlike existing work that either treat the sentence descriptions as a list of nouns (e.g.,~\cite{Cinbis-pami2016, Deselaers-eccv2010, Pandey-iccv2011, Song-nips2014}) or use the entire sentence itself as-is (e.g.,~\cite{Donahue-cvpr2015, Karpathy-cvpr2015, Rohrbach-eccv2016}), we leverage the structure in the descriptions to learn visual groundings on images. 

Specifically, we exploit two types of structural constraints present in the parse tree: \emph{parent-child} (PC) and \emph{sibling-sibling} (SIB) constraints.  The PC (or \emph{inclusive}) constraint enforces the attention mask of any node in the tree to match the \emph{union} of the attention masks of all of its children nodes.  For example, the attention mask of ``a hand with a donut" should encapsulate the masks of both ``a hand'' and ``with a donut" as shown in Fig.~\ref{fig:gtexample}.  The SIB (or \emph{exclusive}) constraint enforces the attention masks of siblings to be exclusive to each other (for example, ``A grey cat" and ``starring at a hand with a donut" in Fig.~\ref{fig:gtexample}).

More formally, the structural loss is defined as $L_{struct} = \lambda_{PC} L_{PC} + \lambda_{SIB} L_{SIB}$, where $\lambda_{PC}$ and $\lambda_{SIB}$ are weights to balance the PC and SIB loss terms, and
\begin{align}
 L_{PC} &= \frac{1}{|P|} \sum_{k \in P} ||A_k - \max_{l \in child(k)} A_l||^2, \label{pcloss} \\
 L_{SIB} &= - \frac{1}{|S|} \sum_{m \in S}  \sum_{pixels \in A} W_m \cdot \log \frac{\max_n A_{m,n}}{\sum_n A_{m,n}}, \label{sibloss}
\end{align}
where $A$ is the attention mask generated for a given phrase. $P$ denotes the set of valid parent nodes, $child(k)$ returns all children nodes of parent node $k$, $S$ is the set of all siblings, $n$ indexes each node in sibling set $m$ (i.e., nodes that are sibling to each other), $\max$ and $\log$ are computed per-pixel, and $(\cdot)$ is element-wise multiplication.

$L_{PC}$ tries to bring the attention mask of a parent node and the \emph{union} mask of all its children nodes to be close to one another, while $L_{SIB}$ introduces competition such that the attention masks of sibling nodes are \emph{exclusive} for every pixel.  $W_m$ is the average per-pixel attention over sibling set $m$: $\frac{1}{n} \sum_n A_{m,n}$.  Its purpose is to enforce stronger exclusivity among sibling attention masks that have high-values in the same pixels.  Without this term, the exclusivity is enforced on \emph{every} pixel equally regardless of whether it is relevant to the current sibling set. This can be problematic when all siblings produce low values for a given pixel (implying that the pixel is irrelevant to the current sibling set), since it will try to undesirably inflate the value for one of the sibling masks.  Empirically, we find this to be the case.

Note that even though we only explicitly enforce the PC constraint between a parent and its immediate children, transitivity ensures that the constraints are carried out through all descendents.  Also, since we only consider a node if it contains at least one noun, each node in a sibling pair (e.g., \emph{NounPhrase-VerbPhrase}, \emph{NounPhrase-PrepositionalPhrase}, or \emph{VerbPhrase-PrepositionalPhrase}) is guaranteed to have its own ``object-of-interest".

\subsubsection{Discriminative loss}

This loss function is used to match the corresponding image-phrase pairs. Given an input image $I_{i}$ and a set of corresponding phrases (both positive and negative ones) $\{P_i^1, P_i^2, ..., P_i^n\}$, we compute the discriminative loss as:
\begin{equation}\label{discloss}
L_{disc} = -Y_i^j \cdot Sigmoid(\phi_V(I_i) \cdot \phi_L(P_i^j)),
\end{equation}
where $Y_i^j \in \{-1, 1\}$ is the indicator variable denoting whether $P_i^j$ is a negative/positive match to $I_i$, and $\phi_V(I)$ and $\phi_L(P)$ denote the visual and language code, respectively.  The positive phrases are those in the parse tree associated with the input image $I_{i}$, while the negative phrases are randomly sampled from those in the parse tree associated with any other image.  This loss tries to bring the visual and language codes for the positive image-phrase pair to be as close as possible, while separating the codes in the negative pair as much as possible.  We measure the affinity between the codes with a dot-product in the semantic embedding space.

%% file: results.tex
\section{Results}

\begin{figure*}[th!]
\centering
\includegraphics[width=1\textwidth]{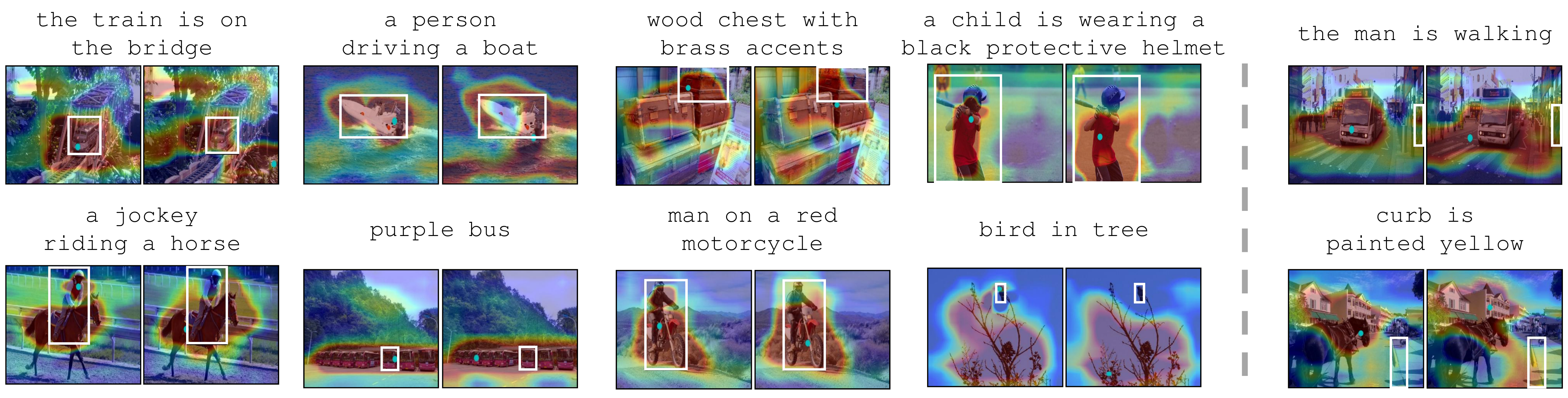}
   \caption{\small{We show qualitative ``pointing game'' results on Visual Genome.  We compare with the baseline model trained without structural constraints (\emph{Disc-only}).  In each image pair, the left is our result and the right is the baseline's result. The ground-truth bounding box is annotated with a white solid line, whereas the maximum point of our prediction is denoted with a cyan dot. The ground-truth phrase associated with each image is shown on top of each image.  The last column shows difficult examples containing small or infrequent objects.}}
\label{fig:visualgenome}
\vspace{-0.1in}
\end{figure*}

\paragraph{Datasets.} We conduct experiments on Visual Genome~\cite{Krishna-arxiv2016} and MS COCO~\cite{Lin-eccv2014} datasets. First, we evaluate on the Visual Genome dataset, which provides caption annotations for image regions.  Since the image regions are annotated with bounding boxes, we use the ``pointing game" \cite{Zhang-eccv2016} to evaluate the capability of our model to visually ground phrases in images. However, since the pointing game evaluation only cares about the maximum point and does not evaluate the full extent of the attention masks produced by our model, we further evaluate our model against the ground-truth category segmentation masks on COCO. For this, we treat category labels as free-form phrases to feed into our language encoder.

\vspace{-10pt}
\paragraph{Baselines.}  We compare to a number of baselines: \emph{Token} is a model that treats the natural language input as a list of object tags during training -- we only take all the leaf nodes with noun POS tags in the parse tree to train the model.  This baseline is meant to represent existing weakly-supervised learning methods that only learn from a list of category labels.  \emph{Disc-only} is a variant of our model that only has the discriminative loss (without the structural loss).
\emph{PC} and \emph{SIB} are each also trained with the discriminative loss but \emph{only} with either the parent-child constraint (Eq.~\ref{pcloss}) or the sibling constraint (Eq.~\ref{sibloss}).  \emph{Ours} is our full model with all loss terms.

\vspace{-10pt}
\paragraph{Implementation details.}  We pre-train the visual (CNN) encoder on ImageNet classification, and pre-train the language (RNN) encoder using the language modeling (next word prediction) task on the combined set of Google Billion Words and COCO captions.  For pre-training, we use the Adam~\cite{Kingma-arxiv2014} solver since it is has been demonstrated to be more robust to sparse updates, which are common in language tasks.  We use SGD with a mini-batch size of 8 images and their associated phrases.  In each batch, the positive samples are images and their corresponding phrases, whereas negative samples are formed by taking an image and sample phrases that do \emph{not} correspond to the image.  We find that fixing the language encoder after pre-training is important to avoid a degenerate solution in which all phrases collapse to almost the same encoding.  For the weights in $L_{struct}$, we set $\lambda_{PC}=0.01$ and $\lambda_{SIB}=0.0001$ based on qualitative examples (see supp.~materials for details on the impact of $\lambda_{PC}$ and $\lambda_{SIB}$).

\subsection{Training on MS COCO}
MS COCO is a large dataset designed for object detection, instance segmentation, and image captioning.  We use the {\tt train2014} and {\tt val2014} sets, which contain 82,783 and 40,504 images, respectively, for training and validation.  We train all variants of our model using images and associated image-level captions on the training set.  We take the trained models and evaluate their localization accuracy using the ``pointing game'' metric and semantic segmentation.
\subsection{Pointing game on Visual Genome}\label{visualgenomeresults}

Visual Genome~\cite{Krishna-arxiv2016} is a recent effort to provide rich annotations on a subset of 328,000 images from MS COCO (it also annotates images from the Flickr30k dataset~\cite{plummer-iccv2015}, which we do not use).  To test the capability of our model in localizing \emph{phrases}, we evaluate on the MS COCO validation set using the region-phrase annotations (i.e., one phrase corresponding to a bounding box in an image) provided by Visual Genome.

Since our approach outputs per-pixel predictions (via the attention mask) instead of bounding boxes, we cannot directly evaluate against the bounding box annotations. This is also why we cannot directly compare with the work of~\cite{Rohrbach-eccv2016}, which outputs bounding boxes rather than segmentation masks. We therefore instead use the \emph{pointing game} evaluation metric~\cite{Zhang-eccv2016}: For each phrase provided in the annotation, we pass the phrase, together with the image to which it associates, through our model and obtain the attention mask on the image.  We then compute the maximum point on the attention mask, and check whether it falls inside the ground-truth bounding box.  If yes, it is counted as a \emph{Hit}; otherwise, it is a \emph{Miss}.  The final accuracy is $\frac{\#Hit}{\#Hit+\#Miss}$.

Table~\ref{table:visualgenomeresults} shows the results.  The model trained with tokens (\emph{Token}) does not perform as well as the model trained using phrases (\emph{Disc-only}), which demonstrates the value of using natural language for image localization.  The parent-child constraint (\emph{PC}) further improves performance over (\emph{Disc-only}), which demonstrates the effectiveness of the parent-child constraint.  While the sibling constraint (\emph{SIB}) performs almost the same as \emph{Disc-only}, when combined with the parent-child constraint, it produces a large boost in performance (\emph{Ours}, our full model), which suggests that the constraints are complementary.  Finally, we also add an extra baseline called \emph{Random} as a sanity check.  For every test image, we select 100 random points and compute the probability of the randomly sampled point falling inside the box.  This baseline clearly performs the worst.

\begin{table}[t!]
\centering
\tabcolsep=0.1cm
    \small
    \begin{tabular}{c || c | c | c | c | c || c }
    \hline
     & Random & Disc-only & Token & PC & SIB & Ours  \\
	 \hline
     Accuracy & 0.115 & 0.230 & 0.222 & 0.236 & 0.231 & \textbf{0.244} \\
    \hline
    \end{tabular}
    \vspace*{0.05in}
    \caption{Localization accuracy as measured by the ``pointing game''~\cite{Zhang-eccv2016} on Visual Genome.   Our model outperforms all baselines, including variants of our method that lack one or more loss terms.  See text for details.}
    \label{table:visualgenomeresults}
    \vspace{-0.1in}
\end{table}

Fig.~\ref{fig:visualgenome} shows qualitative example predictions on the Visual Genome dataset. Overall, our model generates quite accurate masks.  For example, for ``the train is on the bridge" our model accurately pinpoints the train and the bridge whereas the \emph{Disc-only} baseline produces high responses on many irrelevant pixels. A similar thing happens for ``a person driving a boat". Furthermore, in some cases, even though the maximum point of the baseline attention mask falls within the ground-truth bounding box, we can clearly see that the generated mask is not as clean as that of our model; e.g., for ``a child is wearing a black protective helmet" and ``man on a red motorcycle", the baseline model tends to generate leaky masks compared with our results.  This is likely because the baseline model does not have any constraints to exploit other than the discriminative loss, whereas our model explicitly enforces structural priors onto the generated attention masks.

Finally, in the first two rows of Fig.~\ref{fig:compare}, we show different visual groundings corresponding to different input phrases produced by our model for the same input image.  For example, in the first image, our model generates entirely different groundings for ``the clock tower is tall" and ``buildings by the street".  This result demonstrates that our model learns to focus on the right concepts instead of simply computing a language-independent saliency map.

\begin{figure*}[t!]
\centering
\includegraphics[width=1\textwidth]{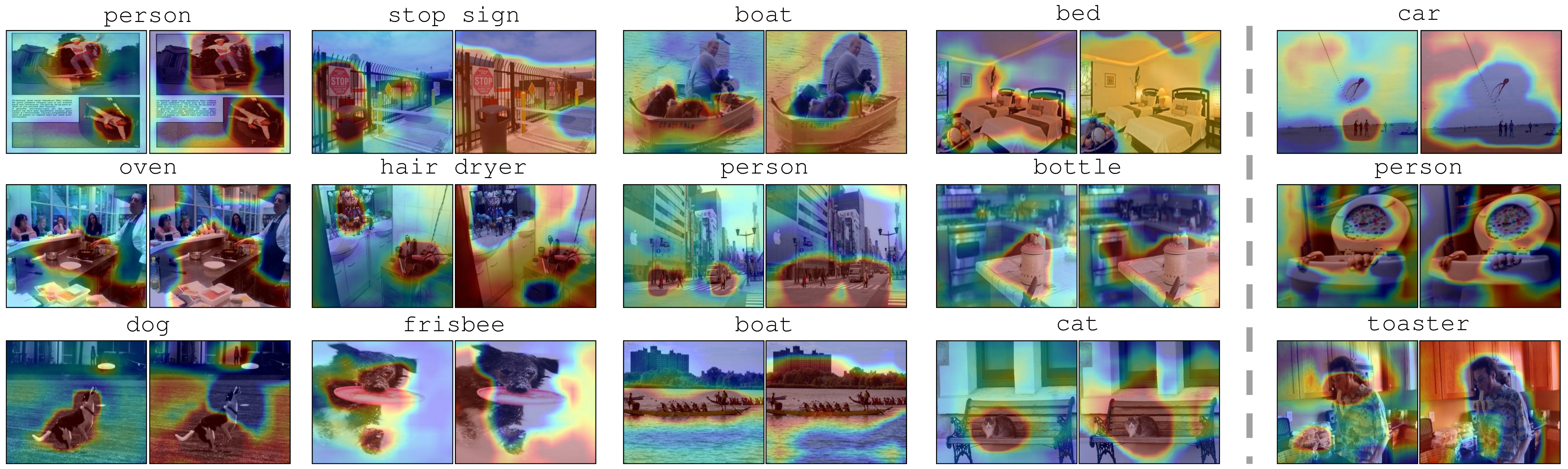}
   \caption{\small{Segmentation results on the MS COCO segmentation task. In each image pair, we show the results of our model on the left and that of the \emph{Disc-only} baseline on the right. The category label is shown on top of each image pair.  The last column shows difficult examples containing small or infrequent objects.  See text for details.}}
\label{fig:coco}
\end{figure*}

\subsection{Semantic segmentation on MS COCO}

Since the ``pointing game'' only evaluates on the maximum prediction point, we next evaluate the full extent of our attention masks through a segmentation task.  For this, we use the MS COCO segmentation annotations~\cite{Lin-eccv2014}.  Note that only category labels are provided (e.g., ``cat") for evaluating segmentation on MS COCO.  In this case, it makes more sense to evaluate our model on semantic segmentation rather than instance segmentation, since our model will only take a single category label as the language input.  We therefore merge instances of the same category into one ground-truth semantic segmentation mask, and evaluate on the validation set (the test set evaluates only instance segmentation and requires submission to a private server).

\begin{table}[t!]
\centering
    \footnotesize
    \begin{tabular}{ c || c | c | c || c }
    \hline
     &  IOU@0.3 & IOU@0.4 & IOU@0.5 & Avg mAP  \\
    \hline
    Disc-only  & 0.302 &  0.199	&  0.110 & 0.203 \\
    PC  & 0.327	& 0.213 & 0.118 & 0.219 \\
    SIB  & 0.316 & 0.203 & 0.114 & 0.211 \\
    Token  & 0.334 & 0.240 &   0.138 & 0.238\\
    \hline
    Ours & \textbf{0.347} & \textbf{0.246} & \textbf{0.159} & \textbf{0.251} \\
    \hline
    \end{tabular}
    \vspace*{0.05in}
    \caption{Segmentation mAP on MS COCO across all 80 categories. Our model produces more accurate segmentations compared to alternate weakly-supervised baselines.}
    \label{table:cocoresults}
    \vspace*{-0.1in}
\end{table}

In order to perform semantic segmentation, we convert our continuous-valued attention masks to binary foreground/background predictions.  For this, we simply binarize it with a threshold, which is set to be the medium value in the predicted attention mask range (i.e., $\theta = \frac{1}{2}(\max(A)-\min(A))$.  We compute the resulting segmented region's prediction score as the average per-pixel attention score within the region.  Finally, we compute the intersection-over-union (IOU) metric for the predicted foreground region against the ground-truth foreground mask.

Table~\ref{table:cocoresults} shows the segmentation results, in term of mean Average Precision (mAP) over all 80 MS COCO categories at different IOU thresholds. Both \emph{PC} and \emph{SIB} provide consistent improvement over \emph{Disc-only} across different IOU thresholds.  This demonstrates that both structural constraints effectively transfer their respective structure from the language domain to the visual domain. Moreover, with our full model \emph{Ours}, which combines both \emph{PC} and \emph{SIB}, mAP is boosted even further. This again shows the complementarity of the parent-child and sibling constraints. Interestingly, the model trained only with noun tokens (\emph{Token}) performs quite well, outperforming both \emph{Disc-only} and \emph{PC}/\emph{SIB}.  This is mainly because for this category semantic segmentation task, the language input is only nouns, and this can favor a model that is also trained using only the noun tokens. Despite this, our full model still outperforms the \emph{Token} baseline and achieves the best performance among all methods. Further, we want to highlight that our model is much more general, as compared to the \emph{Token} baseline; it can localize regions beyond simple objects or noun tokens (e.g., we can localize {\em adjective-noun} or even more complex {\em referring} phrases).

We show example segmentation results in Fig.~\ref{fig:coco}.  One can easily see the improvements brought by the structural constraints. First, our model localizes more accurately (e.g., for ``dog" and ``frisbee"), just like it does on the Visual Genome dataset.  Second,  we again observe the prominent behavior of our model that it tends to generate much cleaner attention masks compared to the \emph{Disc-only} baseline (e.g., ``person", ``cat", ``boat", ``stop sign"). By explicitly encoding the linguistic structural constraints on the visual attention masks, our model is able to obtain more accurate localizations.  The last column shows typical failure cases with small or infrequent objects in the image.

\begin{figure*}[t!]
\centering
\includegraphics[width=1\textwidth]{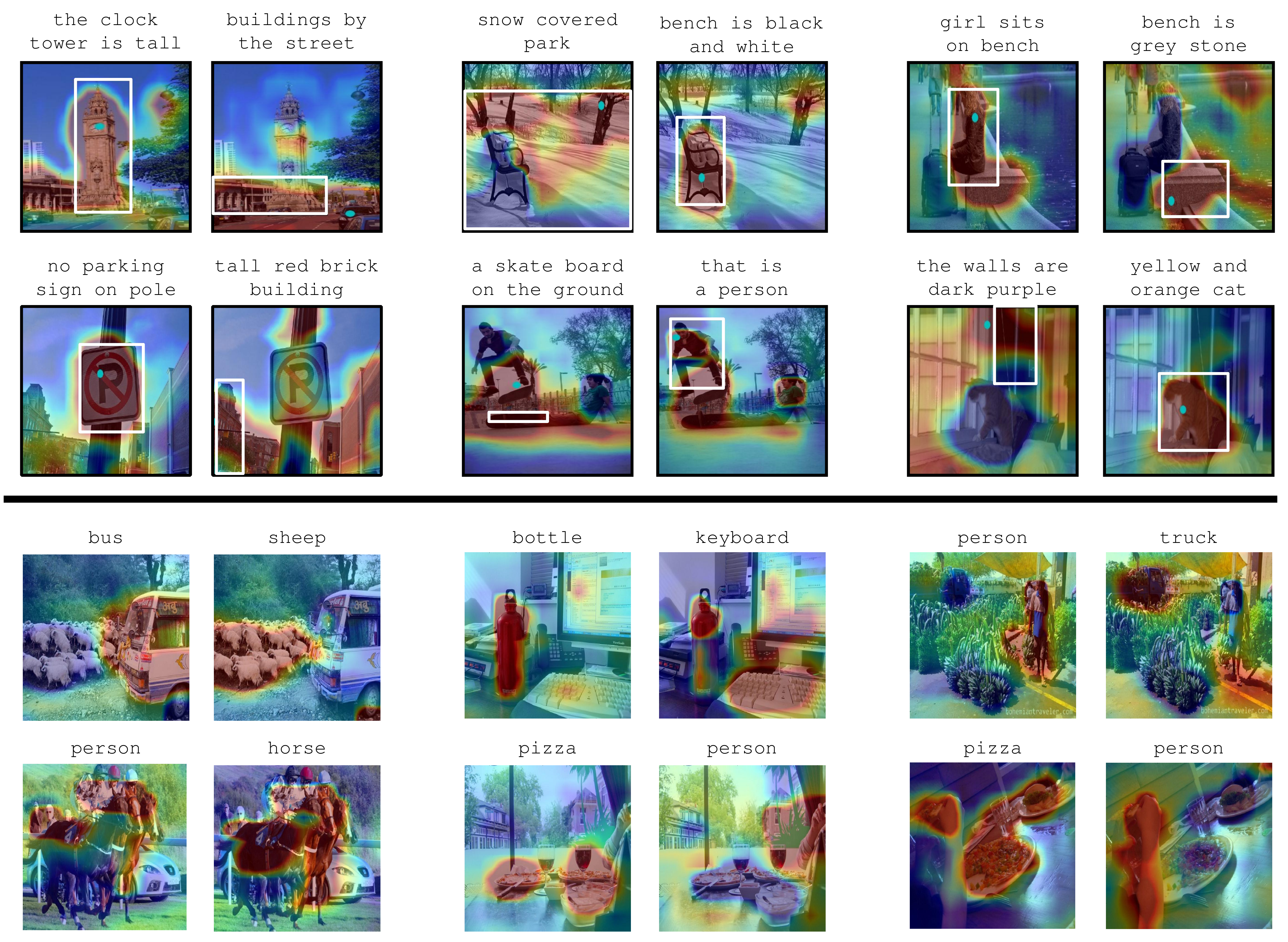}
   \caption{\small{Here we show for the same image, visual groundings generated with different language inputs.  The first two rows are visual grounding results on Visual Genome, with different phrases as language input. The last two rows are visual grounding results on MS COCO, with category labels as language input.  In both cases, our model generates very different visual groundings corresponding to different language input, and correctly focuses on the relevant objects-of-interest.}}
\label{fig:compare}
\vspace{-0.05in}
\end{figure*}

Finally, similar to what we showed on the Visual Genome dataset, the last two rows of Fig.~\ref{fig:compare} show different groundings generated by our model for the same image, given different category labels as language inputs.  Our model is able to detect different objects in the same image.

\subsection{Discussion of failure cases}

While we have shown promising results for this very difficult task of weakly-supervised visual phrase grounding, there are still a few sources of difficulties for our model. 

First, the parse trees produced by the NLP parser can be wrong, which will make the resulting structural constraints insensible.  Empirically, we observe parse tree errors in roughly 20\% of the images.  Another cause of failure is due to the language representation, in that the hidden vectors computed for an entire sentence might not correctly focus on the relevant objects (i.e., entities having corresponding image regions) in the sentence. The third difficulty is that our task is fundamentally weakly-supervised, which in itself limits how well one can do (particularly with respect to full supervision) with \emph{limited training data}. We hypothesis that with a much larger weakly-annotated dataset, we could expect a performance boost on this challenging task.

%% file: conclusion.tex
\section{Conclusion}

We presented a weakly-supervised approach that takes image-level captions, without any explicit region-to-phrase correspondence annotations, and learns to localize arbitrary linguistic phrases in images.  We designed a novel end-to-end deep network with two new structural loss constraints: a parent-child inclusivity constraint and a sibling-sibling exclusivity constraint.  Together, these constraints transfer the structure present in natural language to the visual domain so that the network can learn to produce more accurate visual localizations.  Our experiments on the Microsoft COCO and Visual Genome datasets demonstrate that our approach produces more accurate localizations compared to several baselines that either do not consider any structure or consider only one of the constraints in isolation.